\documentclass{article}

\usepackage{microtype}
\usepackage{graphicx}
\usepackage{subfigure}
\usepackage{booktabs} 

\usepackage{hyperref}

\usepackage[accepted]{icml2024}

\usepackage{amsmath}
\usepackage{amssymb}
\usepackage{mathtools}
\usepackage{amsthm}
\usepackage{threeparttable}

\usepackage{makecell}
\usepackage{multirow}
\usepackage{wrapfig}
\usepackage{colortbl}
\usepackage{tcolorbox}
\usepackage{pifont}
\usepackage{tikz}
\usepackage{xurl}
\usepackage[capitalize,noabbrev]{cleveref}

\theoremstyle{plain}
\newtheorem{theorem}{Theorem}[section]

\theoremstyle{definition}
\newtheorem{definition}[theorem]{Definition}

\theoremstyle{remark}

\definecolor{bgcolor}{rgb}{0.85,0.85,1}

\definecolor{mydarkgreen}{RGB}{39,130,67}
\definecolor{mydarkred}{RGB}{192,25,25}
\newcommand{\green}{\color{mydarkgreen}}
\newcommand{\red}{\color{mydarkred}}
\newcommand{\cmark}{\green\ding{51}}%
\newcommand{\xmark}{\red\ding{55}}%
\newcommand\pcref[1]{(\cref{#1})}

\colorlet{blue}{cyan!60}
\makeatletter
\newcommand{\algcolor}[2]{%
  \hskip-\ALG@thistlm\colorbox{#1}{\parbox{\dimexpr\linewidth-2\fboxsep}{\hskip\ALG@thistlm\relax #2}}%
}
\newcommand{\algemph}[1]{\algcolor{bgcolor}{#1}}
\makeatother
\usepackage[textsize=tiny]{todonotes}

\icmltitlerunning{PrE-Text: Training Language Models on Private Federated Data in the Age of LLMs}

\begin{document}

\twocolumn[
\icmltitle{PrE-Text: Training Language Models on \\ Private Federated Data in the Age of LLMs}



\icmlsetsymbol{equal}{*}

\begin{icmlauthorlist}
\icmlauthor{Charlie Hou}{cmu}
\icmlauthor{Akshat Shrivastava}{meta}
\icmlauthor{Hongyuan Zhan}{meta}
\icmlauthor{Rylan Conway}{meta}
\icmlauthor{Trang Le}{meta} \\
\icmlauthor{Adithya Sagar}{meta}
\icmlauthor{Giulia Fanti}{cmu}
\icmlauthor{Daniel Lazar}{coldrays}
\end{icmlauthorlist}

\icmlaffiliation{cmu}{Department of Electrical and Computer Engineering, Carnegie Mellon University}
\icmlaffiliation{meta}{Meta}
\icmlaffiliation{coldrays}{Coldrays}

\icmlcorrespondingauthor{Charlie Hou}{hou.charlie2@gmail.com}

\icmlkeywords{Machine Learning, ICML}

\vskip 0.3in
]



\printAffiliationsAndNotice{} 

\begin{abstract}
On-device training is currently the most common approach for training machine learning (ML) models on private, distributed user data. Despite this, on-device training has several drawbacks:  (1) most user devices are too small to train large models on-device, (2) on-device training is communication- and computation-intensive, and (3) on-device training can be difficult to debug and deploy.
To address these problems, we propose Private Evolution-Text (PrE-Text), a method for generating differentially private (DP) synthetic textual data.  First, we show that across multiple datasets, training small models (models that fit on user devices) with PrE-Text synthetic data outperforms small models trained on-device under practical privacy regimes ($\epsilon=1.29$, $\epsilon=7.58$). We achieve these results while using  9$\times$ fewer rounds,  6$\times$ less client computation per round, and 100$\times$ less communication per round.
Second, finetuning large models on PrE-Text's DP synthetic data improves large language model (LLM) performance on private data across the same range of privacy budgets. Altogether, these results suggest that training on DP synthetic data can be a better option than training a model on-device on private distributed data. Code is available at \href{https://github.com/houcharlie/PrE-Text}{https://github.com/houcharlie/PrE-Text}.
\end{abstract}

\begin{table*}[t!]
	\centering
	\caption{PrE-Text provides the privacy guarantees of on-device training while (1) being much cheaper in communication and computation, and (2) being much easier to practitioners to deploy in real systems.}\label{table:first}
	\begin{threeparttable}
		\footnotesize\setlength\tabcolsep{5.pt} 
		\begin{tabular}{ c  c  c  c  c  c }
			\toprule[.1em]
			 \begin{tabular}{c}\bf Method \end{tabular} &  \begin{tabular}{c}\bf Train  \\  \bf LLMs? \tnote{\color{blue}(a) } \end{tabular} &\begin{tabular}{c}\bf Is communication \\  \bf cheap? \tnote{\color{blue}(b) }  \end{tabular} & \begin{tabular}{c} \bf  Is computation \\ \bf cheap? \tnote{\color{blue}(c) }  \end{tabular}    & \begin{tabular}{c}\bf Easy to \\ \bf deploy? \tnote{\color{blue}(d) }\end{tabular}  & 
			 \begin{tabular}{c} \bf Privacy-preserving? \end{tabular} \\
			\midrule
			\begin{tabular}{c} Private on-device training \\
   \tiny DP-FedAvg  \citep{mcmahan2017learning} \\
   \tiny DP-FTRL \citep{kairouz2021practical} \end{tabular}  & \xmark & \xmark & \xmark &  \xmark & \cmark\\ 
   
			\hline
                \begin{tabular}{c} Centralized \\ non-private training \end{tabular} & \cmark & \cmark & \cmark & \cmark & \xmark \\
                \hline
			\cellcolor{bgcolor} \begin{tabular}{c}  PrE-Text \\ \centering  \tiny (proposed) \end{tabular} & \cellcolor{bgcolor}\cmark &\cellcolor{bgcolor} \cmark &\cellcolor{bgcolor} \cmark & \cellcolor{bgcolor}\cmark  & \cellcolor{bgcolor} \cmark \\			
			\hline			
		\end{tabular} 
	\begin{tablenotes}
		{  \item [{\color{blue}(a)}] \textbf{Training LLMs.} 
        On-device finetuning of large models (like LLMs) is not feasible because LLMs are too big. PrE-Text allows us to finetune LLMs because the resulting synthetic data is located on-server.  
			\item [{\color{blue}(b)}] \textbf{Communication cost.} 
   PrE-Text required 100x less communication per round (and 9x fewer rounds) in our experiments.
			\item [{\color{blue}(c)}] \textbf{Client computation cost.} 
   PrE-Text required 6x less client computation per round (and 9x fewer rounds) than on-device training in our experiments. This comes at the cost of more server-side computation resources, which is often much less constrained.
			\item[{\color{blue}(d)}] \textbf{Practicality.} PrE-Text produces synthetic data on-server, which allows practitioners to see the training process end-to-end;  this improves debuggability \citep{augenstein2019generative}. Furthermore, the synthetic data can be reused an unlimited number of times without incurring additional privacy cost.
   }
	\end{tablenotes}  		
	\end{threeparttable}
\end{table*}

\section{Introduction}
\label{sec:intro}
In many language applications, such as mobile keyboard autocompletion \citep{mcmahan2017learning}, or instruction-following large language models \citep{yu2024privacy}, training a model on private user data can significantly improve  model performance. 
However, user data is often sensitive, necessitating the use of algorithmic techniques for protecting privacy.  
Federated Learning (FL) \citep{pmlr-v54-mcmahan17a} is a prominent technique that trains models on user devices (we call this on-device training) and aggregates the resulting  model updates at a central server.  Recent works have shown that FL combined with differential privacy (DP) \citep{dwork2006differential}---a combination we refer to as \emph{DP-FL}---can protect privacy while also improving model performance in user applications \citep{mcmahan2017learning, kairouz2021practical, kairouz2021distributed, nguyen2022federated, xu2023learning}.

On-device training or federated learning has several drawbacks. (1) Due to limited on-device storage and computation, client devices cannot be used to train large language models (LLMs) \citep{radford2019language, touvron2023llama}.  
As LLMs become more critical in many use-cases, this becomes more limiting \cite{charles2023towards}. 
(2) On-device training can have high communication and computation costs for clients \citep{cai2022fedadapter}. Indeed, there is a large body of literature studying how to improve the efficiency of on-device training \citep{wang2020tackling, li2020federated, karimireddy2020scaffold, hou2021fedchain, wang2021field,mishchenko2022proxskip,sadiev2022communication,grudzien2023can}.
(3) It is difficult to deploy and debug \citep{augenstein2019generative}, requiring extensive infrastructure investment \citep{The_TensorFlow_Federated_Authors_TensorFlow_Federated_2018, flsim, pytorchios}.

\textbf{An alternative paradigm: Train or finetune on differentially private (DP) synthetic data.}
We propose to have the central server first generate DP synthetic data from private client data, then centrally finetune a pretrained language model on that private synthetic data. 
As in on-device training, clients send DP information to the server; this is used by the server to generate high-quality synthetic data. Unlike on-device training, clients do not need to run training steps for the downstream model. Finetuning on DP synthetic data located on-server (1) eliminates the model size constraints of on-device training, (2) is easier to debug as we can observe the training process without compromising DP, and (3) does not require new training infrastructure, unlike on-device training. Furthermore, DP synthetic data located on server can be reused an unlimited number of times to train many models without incurring additional privacy cost, due to the post-processing property of DP \citep{dwork2006differential}.

Unfortunately, existing techniques for generating DP synthetic language data from federated clients are too low-quality to train or fine-tune a language model \cite{augenstein2019generative}.
We fill this gap by leveraging Private Evolution (PE) \citep{lin2023differentially}, a recent algorithmic breakthrough in DP synthetic data. PE is a framework for generating realistic DP synthetic image data \citep{lin2023differentially}, which achieves high scores in realism metrics like FID \citep{heusel2017gans}. However, \citet{lin2023differentially} do not apply PE to text, nor do they show that training on DP synthetic data is a competitive alternative to direct DP training (DP-FL or DP-SGD \citep{abadi2016deep}) on private data. 
Our work utilizes PE in the natural language setting (which is a nontrivial adaptation) as part of our overall algorithm, then demonstrates through extensive experiments that the resulting synthetic data can produce better models than DP-FL at a fraction of the cost. 

We list our contributions below:

\textbf{(1) PrE-Text (Private Evolution-Text) algorithm.} 
We propose PrE-Text, a new algorithm for DP synthetic text generation.  We build on the following insights: (1) PE must generate variations of samples (e.g., similar images). We adapt this requirement to the language domain by carefully utilizing mask-filling models \citep{devlin2018bert, bart} instead of the diffusion models they used for images. (2) PE alone does not generate enough high-quality synthetic data to effectively finetune an LLM. 
Hence, we add a post-processing phase, in which we use the outputs of PE to seed high-quality LLMs trained on public data to generate more similar text.\footnote{In concurrent work, \citet{xie2024differentially} extend PE to the text domain using similar techniques to ours, developing a new algorithm called Aug-PE. Some differences: (1) Their focus is on generating better DP synthetic text data, whereas ours is on understanding whether DP synthetic data can take the place of on-device learning. To this end, they study the centralized setting, while we study the federated setting. (2) PrE-Text uses the original PE as a subroutine in a larger algorithm, whereas Aug-PE is Private Evolution with core components rewritten.} 
Done carefully, this can generate orders of magnitude more useful training data, greatly aiding generalization (all without incurring more privacy cost, due to the post-processing property of DP \citep{dwork2006differential}).

\textbf{(2) Experimental results.} We produce high-quality DP synthetic language data using PrE-Text, and demonstrate its superiority over other methods in two major settings: 

\textbf{a) Models served on-device.}  
These models are small enough to fit on user devices. We show that in this setting, models trained on synthetic data produced by PrE-Text achieve similar or better performance to models trained on-device (at $\epsilon = 1.29$ and $\epsilon=7.58$, these privacy levels are standard \citep{mcmahan2017learning}), with {\raise.17ex\hbox{$\scriptstyle\sim$}}100$\times$ less communication per round, {\raise.17ex\hbox{$\scriptstyle\sim$}}6$\times$ lower client computation per round, and 9$\times$ fewer rounds. We show that in these important privacy regimes,  PrE-Text outperforms on-device training for training next-token-prediction models.

\begin{figure*}[t!]
    \centering
    \includegraphics[width=\textwidth]{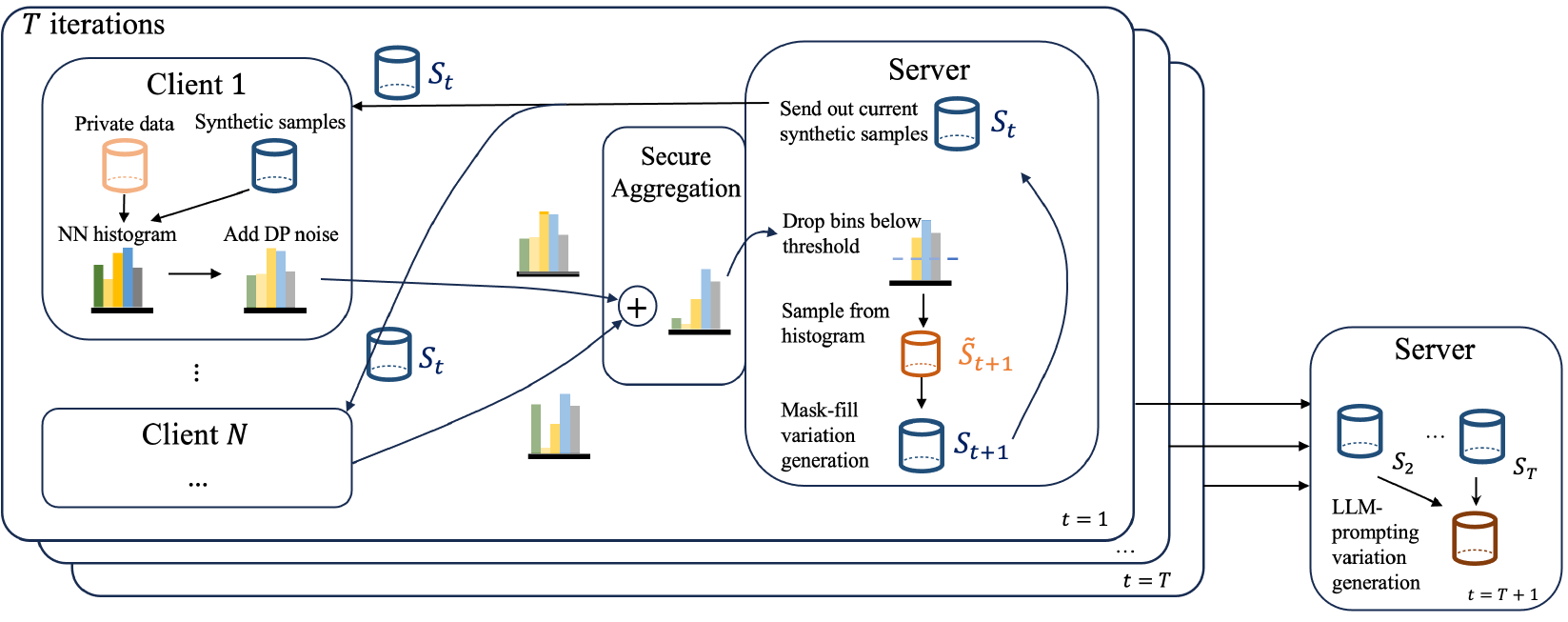}
    \caption{A high-level description of PrE-Text.  PrE-Text consists of two main phases: (1) (iterative) DP synthetic seed collection, (2) (single-shot) synthetic seed expansion.  A detailed description of steps in the diagram is given in \cref{sec: alg description}.}
    \label{fig:alg diagram}
    \end{figure*}

\textbf{b) Models served on-server.} 
These are the models that are too large (in our setting, LLMs) to be served on user devices. We demonstrate that in this setting, {large models finetuned on synthetic data produced by PrE-Text perform better in next-token-prediction than non-finetuned pretrained LLMs}. To the best of our knowledge, we are the first to privately finetune an LLM in the federated setting without requiring the model to be held on user devices. With major LLM providers running out of useful \textit{public} data to train on \citep{runout, runout2, runout3, runout4}, PrE-Text offers a promising path forward. PrE-Text can use \textit{private} client data in a way that is both mindful of user hardware constraints and is also privacy-compliant.

\section{Preliminaries}
In this section, we provide formal definitions for our setting.

\begin{definition}[Neighboring datasets]
    Two federated datasets $X, X'$ are said to be neighboring  (denoted $X\sim X'$) if they differ at most with respect to only one user's data (i.e. $X$ has an extra user compared to $X'$ or vice versa). Note that we consider a \emph{user-level} notion of neighboring datasets \citep{mcmahan2017learning}.
\end{definition}

\begin{definition}[Differential Privacy]
A randomized algorithm $\mathcal{A}$ is $(\epsilon, \delta)$-differentially private (DP) if for any pair of neighboring datasets $X$, $X'$ and for all subsets $E$ of outputs, 
\begin{align}
\text{Pr}[\mathcal{A}(X) \in E] \leq e^{\epsilon}~\text{Pr}[\mathcal{A}(X') \in E] + \delta.
\end{align}
\end{definition}
In this work, we  use the Gaussian Mechanism \citep{dwork2006differential} for DP, which adds Gaussian noise of a specific scale to released statistics. 
The required scale of noise depends on the \emph{sensitivity} of the statistical query we wish to release. When defining the Gaussian noise, we use $I_n$ to represent the identity matrix of size $n \times n$.

\begin{definition}[$\ell_2$ sensitivity \citep{dwork2014analyze}]
Let $g: X \to \mathbb{R}^p$ be a vector-valued function operating on datasets.  Let $X, X'$ be neighboring datasets.  The $\ell_2$-sensitivity of $g$
is defined as 
$\Delta g := \max_{X\sim X'}\|g(X) - g(X')\|_2$.
\end{definition}

\subsection{Problem formulation}
\textbf{Private clients setup.} We consider a setting where a central server wishes to learn a model $\mathcal M$ from $N$ user devices (or, ``clients''). Client $i$ has the language dataset $\mathcal{D}_i$ which consists of $|S_i|$ language samples.  $M$ of these clients, $\mathcal{C}_{\text{test}} \subset [N]$ are considered test clients, and we cannot access their data during synthetic data generation and the training process. The remaining $N-M$ clients, $\mathcal{C}_{\text{train}} \subset [N]$, are considered training clients, and their data can be accessed during model training and/or synthetic data generation. We assume that the client language datasets are drawn from a distribution of possible client datasets $\hat{\mathcal{D}}$, so each $\mathcal{D}_i$ is drawn independently from $\hat{\mathcal{D}}$, $\mathcal{D}_i \sim \hat{\mathcal{D}}$.

We divide the space of models into two: on-device models, which can fit on a client device, and on-server models, which cannot. We assume that large language models (LLMs) and other large foundation models are on-server models. 

\textbf{Server setup.}  The server has access to pretrained LLMs (for example, the LLaMA models in our setting \citep{touvron2023llama}) which were trained only on public data. 

\textbf{Task.} We focus on the language modeling task, where a language model predicts token $s_k$ from the previous tokens $s_0, \cdots, s_{k-1}$ for each text sample. 
The server's final goal is to learn a language model that performs well on next-token-prediction on the private test dataset $\mathcal{C}_{\text{test}}$.

\textbf{Privacy and threat model.} We consider an honest-but-curious threat model \citep{nguyen2022federated}. Using secure aggregation \citep{secagg}, the server does not see individual client uploads, but rather the aggregated upload across clients.  By adding DP noise, clients prevent the server from inferring any single client's data from the aggregated upload (which contains an aggregated amount of noise). The server then aims to learn an $(\epsilon,\delta)$-DP language model (where the notion of DP is user-level distributed DP\footnote{Note that technically, using the Gaussian Mechanism together with secure aggregation requires discretization as secure aggregation uses modular arithmetic \citep{kairouz2021distributed, bagdasaryan2021towards, agarwal2021skellam}. This is an important consideration when practically deploying end-to-end DP applications relying on secure aggregation and the Gaussian mechanism.}). 

\begin{algorithm}[t]
    
    \caption{\texttt{PrE-Text} \label{private evolution}}
    \begin{algorithmic}[1]
    \Statex {\bfseries Input:}
     Number of iterations: $T$\\
     \hspace{\algorithmicindent} Number of generated samples: $N_{\text{syn}}$\\
     \hspace{\algorithmicindent} Initial population: $S_1$ \label{line:initialpop} \\
     {\bfseries Output:} Synthetic data $S_{\text{syn}}$ 
    \For{$t \gets 1 \dots T$} 
        \State \textsc{// see \cref{algo:dphistogram}}
        \State $\text{hist}_t \gets \texttt{Histogram}(S_t)$ \label{line:dp_hist}
        \State $P_t$ $\gets$ $\text{hist}_t / \text{sum}(\text{hist}_t)$
        \State $S_t'$ $\gets$ draw $N_{\text{syn}}$ samples from $P_t$ \label{line:surviving}
        \State \algemph{$S_t \gets \texttt{Variation}(S_t')$ }\label{line:variation}
    \EndFor

      \State  \algemph{$S_{\text{syn}} \gets \texttt{Expand}(\cup_{t=1}^T \texttt{Set}(S_t'))$ \label{line:expand}}
    \State \Return $S_{\text{syn}}$
    \end{algorithmic}
\end{algorithm}

\begin{algorithm}[t]
    \caption{\texttt{Histogram}}
    \begin{algorithmic}[1]
    \State {\bfseries Settings:}  Private samples: Clients $\{C_i\}_{i \in \mathcal{C}_{\text{train}}}$\\
     \hspace{\algorithmicindent} Noise multiplier for histogram: $\sigma$\\
     \hspace{\algorithmicindent} Number of generated samples: $N_{\text{syn}}$\\
     \hspace{\algorithmicindent} Threshold for histogram: $H$ \\
     \hspace{\algorithmicindent} Distance function: $d(\cdot, \cdot)$ 
    \State {\bfseries Input:} Generated samples $S = \{z_i\}_{i=1}^{N_{\text{syn}}}$ \\
     {\bfseries Output:} Nearest neighbors histogram on $S$
    \For{$i \in \mathcal{C}_{\text{train}}$}
        \State {$\text{hist}_i \gets [0, \dots, 0]$ \label{line:fitness}}
        \For{$x_{\text{priv}} \in \mathcal{D}_i$}
            \State $l \gets \text{argmin}_{j \in [N_{\text{syn}}]} d(x_{\text{priv}}, z_j)$ \label{line:dpnn}
            \State $\text{hist}_i[l] \gets \text{hist}_i[l] + 1$
        \EndFor
        \State $\text{hist}_i \gets \text{hist}_i + \mathcal{N}(0, (\sigma^2/|\mathcal{C}_{\text{train}}|) I_{N_{\text{syn}}})$ \label{line:dp}
    \EndFor
    \State $\text{hist} \gets \sum_{i \in \mathcal{C}_{\text{train}}} \text{hist}_i$ \label{line:collect} 
    \State $\text{hist} \gets \text{max}(\text{hist} - H, 0)$ \label{line:threshold} (elementwise subtraction)
    \State \Return \text{hist}
    \end{algorithmic}
    \label{algo:dphistogram}
\end{algorithm}
\section{PrE-Text}
\label{sec: alg description}
The main intuition of PrE-Text (and PE) is that public foundation models should be capable of producing samples that are similar to the private data with some non-negligible probability.  Therefore, to generate DP synthetic data similar to private data, we steer the foundation model (privately) towards the user data in a multi-round process.  Briefly, we make several important changes to the PE algorithm: (1) we adapt it from the image setting to the text setting; (2) we exploit synthetic data from the \emph{entire} PE process, rather than only the last round; and (3) we add a post-processing phase that utilizes the output of PE as seeds for another LLM. We provide pseudocode in \cref{private evolution}, and highlight the new contributed steps in color. We now explain \cref{private evolution}.

\textbf{(1) Population of samples.} We start with an initial population of samples $S_1$ (\cref{line:initialpop}).  These samples can come from many different sources as long as they do not contain private information: for example public samples collected from the internet or samples randomly generated by a public generative foundation model.

\textbf{(2) Clients vote for the best synthetic samples.}  For round $t\geq 1$, we determine which of the generated samples $S_t$ represent the private samples the best.  We send all the generated samples $S_t$ to each client.  Each client counts for each generated sample $s\in S_t$ how many private samples had $s$ as their nearest neighbor in $S_t$.
The higher this count is, the ``better'' a generated sample is.  Thus, each client produces a nearest neighbors histogram with $|S_t|$ entries (\cref{line:fitness}). We determine nearest neighbor according to a distance function $d(y,z) = \|\Phi(y) - \Phi(z)\|_2$, where $\Phi$ is an embedding model.

\textit{Lookahead.} To more accurately assess the closeness of a synthetic sample to a private sample, we amend the distance function from $d(y,z) = \|\Phi(y) - \Phi(z)\|_2$ to $d(y, z) = \|\Phi(y) - \frac{1}{K}\sum_{i=1}^K \Phi(z^i)\|_2$, where $z^0, \dots, z^K$ are $K$ variations of $z$ produced by using \texttt{Variation}. \citet{lin2023differentially} also uses this modification. Instead of sending the actual generated samples directly to all the clients, we send $\frac{1}{K}\sum_{i=1}^K \Phi(z^i)$ to all the clients for every $i$ for their nearest neighbors calculation \pcref{line:dpnn}.

\textit{(a) DP Noise.} In \cref{line:dp} each client adds noise to their nearest neighbors histogram to ensure DP. We compute client-level sensitivity assuming a known upper bound on the number of samples per client (e.g., via thresholding). 

\textit{(b) Federated Secure Aggregation.}  In \cref{line:collect}  we securely aggregate the histograms across the users.  Because the generated samples given to all the clients are the same, we sum the histograms using secure aggregation \citep{secagg} to get an aggregate histogram. 

\textit{(c) Thresholding.} When we generate many samples, the majority of the probability mass of the histogram will be noise.  We improve the signal-to-noise ratio by thresholding the histogram at $H$ in \cref{line:threshold} \citep{lin2023differentially}. 

\textbf{(3) Use votes to choose the surviving samples.} We sample from the nearest neighbors histogram to produce surviving generated samples in \cref{line:surviving}, $S_{t+1}'$.  This new list of generated samples (there may be duplicates) tends to give more representation to generated samples that had more private samples close to them.

\textbf{(4) Produce variations of surviving samples.} We use \texttt{Variation} to generate a variation of the surviving samples $S_{t+1}'$ as the new population of samples (\cref{line:variation}). In PE, this was accomplished using diffusion models, which cannot be used for text.
We instead implement \texttt{Variation}  as follows. For each sample in $S'_t$, we produce a variation of it by masking $\textsc{mask}$\% of the tokens randomly, then filling in those masked tokens in with a masked language model. The resulting sample is then masked and filled-in again. This mask-fill process happens $W_{\text{steps}}$ times before returning the variation. We use RoBERTa-large \citep{roberta} as the masked language model.

\textbf{(5) Making efficient use of iterates.} We make several major modifications to the core Private Evolution algorithm to improve our usage of the iterates. First, \citet{lin2023differentially} use the final $S_T$ (\cref{line:variation}) as the synthetic dataset.  However, $S_T'$ contains more information about the private dataset than $S_T$ because \texttt{Variation} destroys information.  Therefore, we use $S_T'$ instead of $S_T$. Second, we find that $S_t'$ for $t=1,\dots,T$ all have valuable synthetic samples and show significant diversity between iterations.  Therefore, we choose to utilize $\bigcup_{t=1}^T \texttt{Set}(S_t')$ \pcref{line:expand} instead of just $S_T'$.  This more effectively utilizes the privacy budget.

\textbf{(6) Post-processing: Use LLM to expand the DP seed set.} 
PE originally used the final $S_T$ as the target synthetic data. 
However, we find that these samples are not high enough quality to fine-tune an LLM. 
We instead use \texttt{Expand} to generate more samples similar to the synthetic samples $\bigcup_{t=1}^T \texttt{Set}(S_t')$, which we use as our (DP) seed set to prompt the LLM.  \texttt{Expand} utilizes the synthetic data generation capabilities of large language models (LLMs) to generate a larger and more useful synthetic dataset.  Note that by the post-processing property of differential privacy \citep{dwork2006differential}, \texttt{Expand} will not leak any additional privacy.  Next we describe how \texttt{Expand} works.

Inspired by highly successful synthetic text generation \citep{alpaca, wang2022self, honovich2022unnatural, roziere2023code}, we generate synthetic text by using large foundational language models.  We follow a process similar to \citet{honovich2022unnatural}: for each synthetic sample to generate, we randomly choose three text samples to emulate from $\bigcup_{t=1}^T \texttt{Set}(S_t')$ (our DP seed set), and ask the LLM to generate a similar sample.  We use open-source LLaMA-2-7B \citep{touvron2023llama} as our large language model.  We provide the full prompt \pcref{fig:prompt}.

\subsection{Privacy analysis}
As noted in \citet{lin2023differentially}, the only function that utilizes private information is \cref{algo:dphistogram}. The DP histogram (\cref{line:collect}) contains private information.  As in \citet{lin2023differentially}, we use the Gaussian mechanism with constant noise multiplier $\sigma$ each time we receive a histogram. Since this is a Gaussian mechanism, we can use the moments accountant from the Opacus library \citep{opacus}. 
Details  can be found in \cref{app:privacy}.


\section{Experiments}
\label{sec:exp results}

\textbf{Models.}  We use RoBERTa-large \citep{roberta} for mask-filling.  We use all-MiniLM-L6-v2 for text embeddings.  We use DistilGPT2 \citep{sanh2019distilbert} to evaluate our methods. 
Finally, we use LLaMA-2-7B \citep{touvron2023llama} for synthetic seed expansion.

\textbf{Datasets.} We produce three federated private datasets from the c4-English \citep{c4} (c4-en): \textsc{Jobs}, \textsc{Forums}, and \textsc{Microblog}, which are subsets of c4-en. In these datasets, the federated datasets are uniformly randomly partitioned among clients. We also produce another federated private dataset \textsc{Code}, a question-and-answer dataset focused on coding and technical topics. For all training datasets, there are 1250 clients. The evaluation sets are created from a held-out portion of the data. For the initial population used in PrE-Text, we use a subset of c4-en that is not part of any of the private datasets. More details on the datasets are provided in \cref{app:dataset}. 

Note that many LLMs do not document what datasets were used in their pretraining, which makes it difficult to prevent contamination. Even text released after the release of the model may be contaminated, as it may have been AI-generated. We used the most recent large-scale dataset we could find (though it was released before the release of LLaMA-2) that is (a) compliant with terms of service (many sources of recent text data have closed their APIs for ML training) and (b) readily accessible. Systematically detecting dataset contamination is an important open problem in LLM research \citep{gunasekar2023textbooks}. 

\textbf{Task.} We focus on the language modeling task, and report evaluation loss (cross-entropy) and accuracy. We consider two experimental settings: (small) models stored on-device, and (large) models stored on-server.

\textbf{Baselines.} We compare PrE-Text to several baselines.

\textit{(1) $\epsilon=0$ baselines:} Our first two baselines, \textbf{c4-only} and \textbf{Expand-only}, give lower bounds on performance by not using the private data at all, and relying only on public data. 
c4-only is a DistilGPT2 model finetuned on a subset of c4-en that came from website sources not represented in any of the private data. As \citet{xu2023federated} found, finetuning on c4-en improved privacy-utility tradeoff greatly in next-token prediction, so this is an important baseline. Expand-only is a DistilGPT2 model finetuned on the subset of c4-en used in the c4-only baseline, expanded to 2 million samples using \texttt{expand}. 

\textit{(2) $\epsilon=\infty$ baselines:} Our next baseline provides an upper bound on model performance, obtained by ignoring privacy constraints. To this end, we evaluate \textbf{Expand-private}. 
This is a DistilGPT2 model finetuned on (1) the subset of c4-en used in c4-only baseline, and (2) the private dataset expanded to 2 million samples using \texttt{expand}. We found that this performs better than a model finetuned on only the private dataset itself.

\textit{(3) On-device baselines:} We next evaluate two representative on-device training baselines that use DP optimization to provide a privacy guarantee: \textbf{DP-FedAvg} \citep{mcmahan2017learning} and \textbf{DP-FTRL} \citep{kairouz2021practical}, which are two of the most widely-used algorithms for private on-device training in practice. Specifically, we first finetune DistilGPT2 on the subset of c4-en used in the c4-only baseline, and then finetune it further using DP-FedAvg or DP-FTRL (which are on-device training methods). We use the DP-FTRL-TreeRestart variant of DP-FTRL, as we consider full participation in each communication round.

\textit{(4) Text-to-text privatization baseline:} We finally compare PrE-Text against \textbf{DP-Prompt}, a different approach for generating DP synthetic data with paraphrasing. In this approach, clients hold an LLM on-device and release privacy-preserving paraphrases of their text directly to the server. The representative method we use here is DP-Prompt \citep{utpala-etal-2023-locally}. We use the same prompt and model (flan-t5-3b) as \citet{utpala-etal-2023-locally}. Note that these methods cannot take advantage of secure aggregation (text cannot be summed) which necessitates adding much more noise to the privatized text to guarantee user-level privacy. We first finetune a DistilGPT2 model on a subset of c4-en used in the c4-only baseline, and then finetune it further on the privatized text received by the server.
\subsection{Models stored on-device}
\label{sec:ondevice details}
\label{sec:ondevice}
We consider a setting where users do not send data directly to a server-side LLM and instead keep a small model on-device for inference.  We use DistilGPT2 as our representative example of a small model, with 82M parameters. 

\textbf{Experimental setup.}  For \textbf{PrE-Text}, we generate a synthetic dataset of 2 million samples. We first finetune DistilGPT2 on the subset of c4-en used in the c4-only baseline, and then finetune it further on the synthetic dataset generated by PrE-Text. We compare against all the baselines mentioned at the beginning of the section. For more details on how we instantiated each method including hyperparameter grids, see \cref{app:exp details}.

\begin{table*}[t]
    \centering
    \caption{We compare the next token prediction accuracies (higher is better) achieved by PrE-Text against DP-Prompt/DP-FedAvg/DP-FTRL, under $(1.29, 3 \times 10^{-6})$-DP and $(7.58, 3 \times 10^{-6})$-DP. We also compare these methods against baselines with $\epsilon=0, \epsilon=\infty$. We see that PrE-Text outperforms the alternatives at $\epsilon=1.29$ and $\epsilon=7.58$. The error bars are stderr.}
    \begin{tabular}{c | c || c | c | c | c}
      \toprule
      Method & Privacy & \textsc{Jobs} ($\uparrow$)  & \textsc{Forums}($\uparrow$)  & \textsc{Microblog} ($\uparrow$)  & \textsc{Code} ($\uparrow$)\\
      \midrule
      \begin{tabular}{c}
     c4-only \\ Expand-only
      \end{tabular} &
      $\epsilon = 0$ & \begin{tabular}{c}
          0.695 $\pm$ 0.000 \\
          0.702 $\pm$ 0.000
      \end{tabular}
      & \begin{tabular}{c}
      0.650 $\pm$ 0.000\\
      0.661 $\pm$ 0.000 
      \end{tabular} &
      \begin{tabular}{c}
      0.658 $\pm$ 0.000 \\
      0.669 $\pm$ 0.000
      \end{tabular} &
      \begin{tabular}{c}
      0.621 $\pm$ 0.000 \\
      0.633 $\pm$ 0.000
      \end{tabular} \\
      \midrule
      \begin{tabular}{c}
     DP-Prompt \\ DP-FedAvg \\ DP-FTRL \\ PrE-Text
      \end{tabular} &
      $\epsilon = 1.29$ & \begin{tabular}{c}
              0.673 $\pm$ 0.000 \\
          0.701 $\pm$ 0.000 \\
            0.703 $\pm$ 0.000 \\
          \textbf{0.718 $\pm$ 0.000}
      \end{tabular}
      & \begin{tabular}{c}
            0.636 $\pm$ 0.000\\
      0.663 $\pm$ 0.000\\
            0.665 $\pm$ 0.000\\
      \textbf{0.672 $\pm$ 0.000} 
      \end{tabular} &
      \begin{tabular}{c}
            0.642 $\pm$ 0.000\\
      0.665 $\pm$ 0.000 \\
            0.667 $\pm$ 0.000\\
      \textbf{0.680 $\pm$ 0.000}
      \end{tabular} &
      \begin{tabular}{c}
            0.607 $\pm$ 0.000\\
      0.636 $\pm$ 0.000 \\
    0.645 $\pm$ 0.000\\
      \textbf{0.650 $\pm$ 0.001}
      \end{tabular} \\
     \midrule
     
      \begin{tabular}{c}
      DP-Prompt \\ DP-FedAvg \\ DP-FTRL \\PrE-Text
      \end{tabular} &
      $\epsilon = 7.58$ & \begin{tabular}{c}
            0.672 $\pm$ 0.000\\
          0.703 $\pm$ 0.000 \\
          0.704 $\pm$ 0.000\\
         \textbf{0.721 $\pm$ 0.000}
      \end{tabular}
      & \begin{tabular}{c}
     0.637 $\pm$ 0.000 \\
      0.666 $\pm$ 0.000\\
      0.667 $\pm$ 0.000 \\
      \textbf{0.673 $\pm$ 0.000} 
      \end{tabular} &
      \begin{tabular}{c}
      0.642 $\pm$ 0.000\\
      0.666 $\pm$ 0.000 \\
      0.668 $\pm$ 0.000\\
      \textbf{0.680 $\pm$ 0.000}
      \end{tabular} &
      \begin{tabular}{c}
            0.607 $\pm$ 0.000\\
      0.637 $\pm$ 0.000 \\
      0.646 $\pm$ 0.000\\
      \textbf{0.656 $\pm$ 0.000}
      \end{tabular} \\
      \midrule

      Expand-private &
      $\epsilon = \infty$ & 
     0.730 $\pm$ 0.000 & 0.684 $\pm$ 0.000& 0.689 $\pm$ 0.000& 0.680 $\pm$ 0.000\\
      
    \bottomrule
    \end{tabular}
    \label{tab:fl}
  \end{table*}

\begin{table*}[t]
    \centering
    \caption{We compare the next token prediction (cross-entropy) losses (lower is better) achieved by PrE-Text against DP-Prompt/DP-FedAvg/DP-FTRL, under $(1.29, 3 \times 10^{-6})$-DP and $(7.58, 3 \times 10^{-6})$-DP. We also compare these methods against baselines with $\epsilon=0, \epsilon=\infty$. We see that PrE-Text outperforms the alternatives at $\epsilon=1.29$ and $\epsilon=7.58$. The error bars are stderr.}
    \begin{tabular}{c | c || c | c | c | c}
      \toprule
      Method & Privacy & \textsc{Jobs} ($\downarrow$)  & \textsc{Forums}($\downarrow$)  & \textsc{Microblog} ($\downarrow$)  & \textsc{Code} ($\downarrow$)\\
      \midrule
      \begin{tabular}{c}
     c4-only \\ Expand-only
      \end{tabular} &
      $\epsilon = 0$ & \begin{tabular}{c}
          1.781 $\pm$ 0.004 \\
          1.611 $\pm$ 0.000
      \end{tabular}
      & \begin{tabular}{c}
      2.154 $\pm$ 0.007\\
      1.883 $\pm$ 0.000 
      \end{tabular} &
      \begin{tabular}{c}
      2.103 $\pm$ 0.004 \\
      1.858 $\pm$ 0.000
      \end{tabular} &
      \begin{tabular}{c}
      2.525 $\pm$ 0.000 \\
      2.195 $\pm$ 0.000
      \end{tabular} \\
      \midrule
      \begin{tabular}{c}
     DP-Prompt \\ DP-FedAvg \\ DP-FTRL \\ PrE-Text
      \end{tabular} &
      $\epsilon = 1.29$ & \begin{tabular}{c}
              1.799 $\pm$ 0.003 \\
          1.644 $\pm$ 0.000 \\
            1.594 $\pm$ 0.000 \\
          \textbf{1.482 $\pm$ 0.001}
      \end{tabular}
      & \begin{tabular}{c}
            2.049 $\pm$ 0.001 \\
      1.888 $\pm$ 0.000 \\
            1.850 $\pm$ 0.000 \\
      \textbf{1.779 $\pm$ 0.001}
      \end{tabular} &
      \begin{tabular}{c}
            2.055 $\pm$ 0.000 \\
      1.916 $\pm$ 0.000 \\
            1.874 $\pm$ 0.000 \\
      \textbf{1.770 $\pm$ 0.000}
      \end{tabular} &
      \begin{tabular}{c}
            2.285 $\pm$ 0.004\\
      2.149 $\pm$ 0.001 \\
    2.032 $\pm$ 0.000\\
      \textbf{1.998 $\pm$ 0.009}
      \end{tabular} \\
     \midrule
     
      \begin{tabular}{c}
      DP-Prompt \\ DP-FedAvg \\ DP-FTRL \\PrE-Text
      \end{tabular} &
      $\epsilon = 7.58$ & \begin{tabular}{c}
            1.801 $\pm$ 0.002\\
          1.598 $\pm$ 0.000 \\
          1.589 $\pm$ 0.000\\
         \textbf{1.456 $\pm$ 0.003} 
      \end{tabular}
      & \begin{tabular}{c}
     2.047 $\pm$ 0.000 \\
      1.854 $\pm$ 0.000 \\
      1.845 $\pm$ 0.000 \\
      \textbf{1.773 $\pm$ 0.001}
      \end{tabular} &
      \begin{tabular}{c}
      2.055 $\pm$ 0.001\\
      1.879 $\pm$ 0.000 \\
      1.867 $\pm$ 0.000 \\
      \textbf{1.771 $\pm$ 0.000}
      \end{tabular} &
      \begin{tabular}{c}
            2.287 $\pm$ 0.000\\
      2.141 $\pm$ 0.000 \\
      2.027 $\pm$ 0.000\\
      \textbf{1.927 $\pm$ 0.002}
      \end{tabular} \\
      \midrule

      Expand-private &
      $\epsilon = \infty$ & 
     1.374 $\pm$ 0.002 & 1.682 $\pm$ 0.001 & 1.668 $\pm$ 0.001 & 1.677 $\pm$ 0.001\\
      
    \bottomrule
    \end{tabular}
    \label{tab:fl loss}
  \end{table*}

\begin{table*}[t]
    \centering
    \caption{Next token prediction accuracy on private data for LLaMA-2-7B finetuned on PrE-Text synthetic data (under $(7.58, 3 \times 10^{-6})$-DP and $(1.29, 3 \times 10^{-6})$-DP). The finetuned model outperforms non-finetuned LLaMA-2-7B  performance across four datasets. The error bars are stderr.}
    \begin{tabular}{c | c || c | c | c | c}
      \toprule
      Method & Privacy &  \textsc{Jobs} ($\uparrow$)  & \textsc{Forums}($\uparrow$)  & \textsc{Microblog} ($\uparrow$) & \textsc{Code} ($\uparrow$) \\
      \midrule
      Non-finetuned & $\epsilon=0$ & 0.458 $\pm$ 0.000 & 0.466 $\pm$ 0.000 & 0.473 $\pm$ 0.000 & 0.460 $\pm$ 0.000 \\
      PrE-Text  & $\epsilon=1.29$ & 0.522 $\pm$ 0.000 & 0.479 $\pm$ 0.001  & 
       0.484 $\pm$ 0.000 &  0.475 $\pm$ 0.000  \\
      PrE-Text  & $\epsilon=7.58$ & 0.523 $\pm$ 0.000  & 
      0.480 $\pm$ 0.000 & 0.483 $\pm$ 0.000 & 0.476 $\pm$ 0.000   \\
      Expand-private & $\epsilon=\infty$ & 0.526 $\pm$ 0.000 & 0.484 $\pm$ 0.000 & 0.488 $\pm$ 0.001 & 0.479 $\pm$ 0.000 \\
    \bottomrule
    \end{tabular}
    \label{tab:zeroshot accuracy}
  \end{table*}

\begin{table*}[t]
    \centering
    \caption{
    Next token prediction cross-entropy loss on private data for LLaMA-2-7B finetuned on PrE-Text synthetic data (under $(7.58, 3 \times 10^{-6})$-DP and $(1.29, 3 \times 10^{-6})$-DP). The finetuned model outperforms non-finetuned LLaMA-2-7B  performance across four datasets. The error bars are stderr.}
    \begin{tabular}{c | c || c | c | c | c}
      \toprule
      Method & Privacy &  \textsc{Jobs} ($\downarrow$)  & \textsc{Forums}($\downarrow$)  & \textsc{Microblog} ($\downarrow$) & \textsc{Code} ($\downarrow$) \\
      \midrule
      Non-finetuned & $\epsilon=0$ & 3.252$\pm$ 0.000 & 3.250 $\pm$ 0.000 & 3.220 $\pm$ 0.000 & 3.218 $\pm$ 0.000 \\
      PrE-Text  & $\epsilon=1.29$ & 2.811 $\pm$ 0.000 & 3.177 $\pm$ 0.004  & 
       3.133 $\pm$ 0.002 &  3.145 $\pm$ 0.000  \\
      PrE-Text  & $\epsilon=7.58$ & 2.795 $\pm$ 0.008  & 
      3.164 $\pm$ 0.005 & 3.139 $\pm$ 0.008 & 3.135 $\pm$ 0.000   \\
      Expand-private & $\epsilon=\infty$ & 2.746 $\pm$ 0.002 & 3.085 $\pm$ 0.001 & 3.088 $\pm$ 0.001 & 3.054 $\pm$ 0.000 \\
    \bottomrule
    \end{tabular}
    \label{tab:zeroshot}
  \end{table*}

\textbf{Results.}
We present our results in \cref{tab:fl} and \cref{tab:fl loss}.
We find that PrE-Text outperforms all other baselines at $\epsilon = 1.29$ and $\epsilon = 7.58$. As a sanity check, it also performs better than the $\epsilon=0$ baseline and worse than the $\epsilon=\infty$ baseline. The results show that in practical privacy regimes, PrE-Text outperforms private on-device training. As synthetic data generation methods improve (for example, better prompts and models for \texttt{expand}), synthetic data generation may continue to greatly improve as a strategy for learning from private federated datasets.

\textbf{Efficiency differences.}  
We compare the efficiency of DP-FL vs PrE-Text in our experimental setup, demonstrating that PrE-Text is much more efficient in terms of computation and communication in our setting.

\textit{(1) Communication cost.}  DP-FL requires each client to download and upload the model (DistilGPT2) to/from the server. DistilGPT2 has a size of 82M parameters, which means clients are downloading and uploading 82M floats each round. On the other hand, PrE-Text requires clients to download at most 2048 embedding vectors of size 384 representing the synthetic samples each round. This means each client downloads around 800K floats. On upload, clients upload at most 2048 floats (the size of the histogram). So client download cost is 100$\times$ cheaper with PrE-Text per round, and upload cost is 41000$\times$ cheaper with PrE-Text per round. In addition, PrE-Text uses 9$\times$ fewer rounds than the on-device baselines. Conservatively, this is at least a 100$\times$ improvement in communication cost per round when using PrE-Text (possibly more, given that upload speeds can be 15$\times$ slower than download \citep{speedtest.net}). The embedding model we use, miniLM-L6-v2 \citep{reimers-2019-sentence-bert}, is itself only 10M floats, and only needs to be downloaded once per user, and can be done offline.

\textit{(2) Client computation cost.}  PrE-Text is also much more computationally efficient for clients.  We assume that user devices only have access to CPU computation, as most smartphones do not have GPUs (or have fairly limited GPUs).  When tested on a VM with five Intel(R) Xeon(R) Gold 6248 CPU @ 2.50GHz, training with DistilGPT2 requires 3 seconds per sample to train while the client computation for PrE-Text (nearest neighbors calculation and client embedding generation) requires less than 0.5 seconds per sample. This computation gain comes from the fact that clients perform inference (not training), which requires fewer operations and can be sped up \citep{kwon2023efficient, reimers-2019-sentence-bert}. This gives at least a 6$\times$ improvement in computation cost per round for clients with PrE-Text. In addition, PrE-Text uses 9$\times$ fewer rounds than the on-device baselines. This comes at the cost of more server-side compute in our experiments. However, using server resources is often more acceptable than using client resources.
\begin{table*}[]
\centering
\begin{tabular}{ c|c c|c|c}
\toprule
\multirow{2}{*}{\textbf{Method}} & \multicolumn{2}{l|}{\begin{tabular}[c]{@{}l@{}}\textbf{Per-Round Communication} \\ (Number of floats transmitted)\end{tabular}} & \multirow{2}{*}{\begin{tabular}[c]{@{}c@{}}\textbf{Per-sample Runtime}\\ (s)\end{tabular}} & \multirow{2}{*}{\begin{tabular}[c]{@{}c@{}}\textbf{Number of rounds}\\ \end{tabular}} \\ \cline{2-3}
                                 & \multicolumn{1}{l|}{\textit{Upload}}                               & \textit{Download}                              &                                                                                                   \\ \midrule
DP-FL                            & \multicolumn{1}{l|}{82M}                                           & 82M                                            & 3 & 100                                                                                 \\
PrE-Text                         & \multicolumn{1}{l|}{2048}                                          & 800K                                           & 0.5  & 11                                                                                            \\ \midrule
Reduction                        & \multicolumn{1}{l|}{41,000$\times$}                                       & 100$\times$                                            & 6$\times$    & 9$\times$                                                                                          \\ \bottomrule
\end{tabular}
\label{tab:efficiency}
\caption{Communication and runtime cost of PrE-Text vs DP-FL. We find that PrE-Text achieves at least a 100$\times$ reduction in per-round communication, a $6\times$ reduction in per-round runtime, and a 9$\times$ reduction in the number of rounds in our experiments. These costs are the same for all datasets and values of $\epsilon$ in our evaluation, for private clients with 8 samples each. The number of synthetic samples sent to each client for PrE-Text is 2048.
}
\end{table*}
\subsection{Models stored on-server}

We next consider the setting where models are stored and trained on-server (as opposed to on-device). This setting arises when the model is too large to fit on-device, for instance. 
On-device training  and inference are infeasible in this setting. We instead obtain a DP synthetic dataset from the private federated training set and then finetune the server-side LLM on this synthetic dataset.

\textbf{Experimental details.}  We use the same settings for PrE-text as in \cref{sec:ondevice}, except we only expand to 50000 samples (due to computational constraints of finetuning LLMs).  We use LLaMA-2-7B as our evaluation model instead of DistilGPT2.  We compare against the relevant and competitive baselines. For the non-finetuned (on private data) baseline, we simply report the evaluation loss on the private datasets for LLaMA-2-7B.  To evaluate our proposed alternative, we finetune LLaMA-2-7B on the synthetic dataset produced by PrE-Text for one epoch with LoRA finetuning (with rank 4, $\alpha=8$, applied to all the projection matrices in LLaMA-2-7B) with the AdamW optimizer at a learning rate of 0.0002 and a batch size of 512. For the Expand-private ($\epsilon=\infty$) baseline we finetune on 50000 samples expanded from the private train set.
The on-device baselines are not appropriate for this setting. We also do not compare against DP-Prompt in this setting because it performed very poorly in \cref{sec:ondevice}, even worse than the $\epsilon=0$ baseline. 

\textbf{Results.} In \cref{tab:zeroshot accuracy} and \cref{tab:zeroshot} we observe that LLaMA-2-7B finetuned on PrE-Text DP synthetic data outperforms zero-shot LLaMA-2-7B. To the best of our knowledge, we are the first to demonstrate a way to privately finetune a large language model (that cannot fit on-device) in the federated setting. With major LLM providers running out of useful \textit{public} data \citep{runout, runout2, runout3, runout4} for training, our proof-of-concept shows a promising new path forward: using \textit{private} client data in a privacy-compliant and resource-conscious way.

\subsection{Performance Scaling by Dataset Size}
\begin{figure}
    \centering
    \includegraphics[width=0.85\columnwidth]{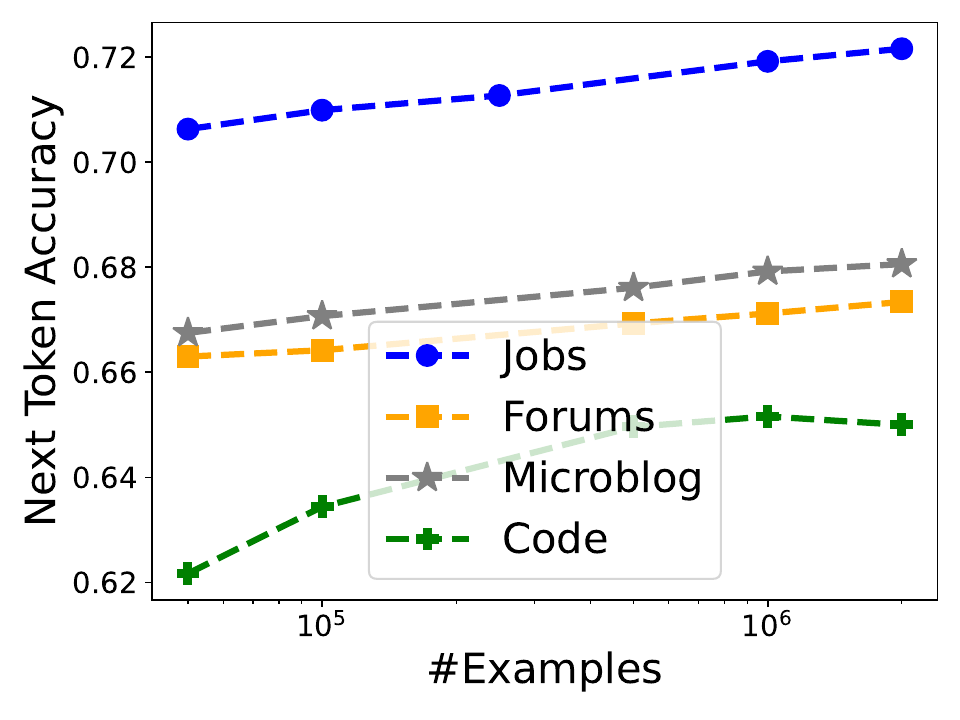}
    \caption{Next-token prediction accuracy for PrE-Text as we vary the number of synthetic examples generated by the \texttt{Expand} part of the algorithm. We find that increasing the number of synthetic examples across several orders of magnitude improves the accuracy of the downstream model (DistilGPT2) roughly log-linearly, though the growth peaks on \textsc{Code} after 1M samples.
    }
    \label{fig: scaling}
\end{figure}

In this subsection, we investigate how the performance of PrE-Text scales as we change the number of samples generated in \texttt{Expand}. We finetune DistilGPT2 on DP synthetic datasets of sizes ranging from 50000 to 2 million, at $\epsilon=7.58$ across the four private datasets  \textsc{Jobs},  \textsc{Forums}, \textsc{Microblog}, \textsc{Code}. All the experimental details are the same as in the on-device PrE-Text experiment, except we scale the batch size according to the fraction of synthetic data we use over the full amount (2 million). For example, at a synthetic data size of 50000, we use [0.0025 $\times$ the original batch size] as the batch size. As shown in \cref{fig: scaling}, we find that like in \citep{honovich2022unnatural}, the quality of the downstream model mostly improves log-linearly with the amount of synthetic data. However, in the case of \textsc{Code}, the performance of the downstream model starts getting diminishing returns at around 1M samples, and even decreases in performance at 2M samples. The best number of synthetic samples for downstream may depend on the dataset. 


\section{Related Work}


\textbf{DP Federated Learning.} Differentially Private Federated Learning (FL) is a widely-used approach for learning ML models from distributed private data \citep{mcmahan2017learning,kairouz2021advances}. In DP-FL, model weights are sent to users who train the model locally on their private data; private local model updates are then collected at a central server. 
Researchers have many techniques for improving privacy-utility tradeoffs in DP-FL. Some include shuffle-based privacy amplification  \cite{secagg,girgis2021shuffled,agarwal2018cpsgd}, pretraining on public datasets \cite{xu2023federated}, private selection of the best pretraining datasets \cite{hou2023privately,gu2022choosing}, and DP-FL methods that do not rely on uniform sampling/shuffling \citep{kairouz2021practical}.

Today, growing efforts study how to train larger models on client data.  \citet{charles2022federated} propose to have users optimize slices of large models, though they have not demonstrated the approach on models larger than shallow logistic regression and convolutional neural network models.  \citet{collins2023profit, cai2022fedadapter, zhang2023towards, zhao2022reduce, guo2023promptfl} only tune sub-components of the models in the federated setting to reduce client computational burden and communication, but this still requires having clients store and perform inference with large models on their device.  \citet{zhang2023gpt} use foundation models to produce synthetic data to pretrain smaller models.  None of these methods consider privacy.



\textbf{Synthetic Data.} Synthetic data is increasingly being used to train language models \citep{alpaca, wang2022self, honovich2022unnatural}.  Common approaches involve carefully designing prompts to ChatGPT \citep{radford2019language} to generate synthetic training data for another open source model, e.g. LLaMA \citep{touvron2023llama}, to replicate ChatGPT behavior.
In these works, the synthetic data is used solely to enhance final model utility, and does not satisfy any formal privacy guarantees. In the image setting, useful synthetic data is often produced using dataset distillation \citep{wang2018dataset, zhao2020dataset, cazenavette2022dataset}. This approach has been adapted to the federated setting \citep{song2023federated}.

\textbf{DP Synthetic Data.} Much of the work on producing \emph{private} (i.e., DP) synthetic data from deep generative models is in the image domain \citep{lin2021privacy, cao2021don,dockhorn2022differentially,chen2022fedtune}. 
Common approaches involve training a generative adversarial network (GAN) or diffusion model in a differentially private way, e.g., via DP-SGD, a DP version of stochastic gradient descent (SGD) \citep{abadi2016deep}.  \citet{tang2023privacy} generate DP synthetic text in the federated setting, but their method requires users to send private data to ChatGPT (which is located on-server).  
Such actions are not allowed under our threat model, where the central server is not trusted to hold private data. 
In concurrent work, \citep{xie2024differentially} propose a  private evolution algorithm for DP synthetic text data. Their method is similar to ours (we summarize differences in \cref{sec:intro}); the most important difference is that our focus is on understanding the relation between synthetic data and on device training, while their work aims to improve the quality of DP synthetic text data more generally.
Also in the text domain, \citep{li2021large, yu2021differentially} demonstrate that it is possible to finetune pretrained large language models in a central-DP manner (i.e., the private data is available to the model developer).

Recently, 
several papers (some of them concurrent to ours) have proposed to finetune LLMs on DP synthetic data in the central DP setting \citep{yue2022synthetic, kurakin2023harnessing,yu2024privacy, ding2024delving}. 
In the private federated setting, work by \citet{wang2023can, wu2024prompt}
 also considers using LLM synthetic data. They propose to filter LLM synthetic data to samples that are relevant to the private data. The model is subsequently finetuned using DP-FL on-device. 
Their work shows a substantial improvement in next-word prediction accuracy compared to vanilla pre-training with DP-FL finetuning. In contrast, our work uses DP synthetic data to replace the DP-FL step itself, while keeping pre-training untouched. These methods are complementary and could potentially be combined.
Overall, the literature suggests that DP synthetic data may be useful for training models from private, client data.



\section{Conclusion}
We propose PrE-Text, a method for generating privacy-preserving synthetic text from user devices that (1) surpasses federated learning approaches across privacy settings and (2) allow us to privately finetune LLMs on distributed user data. As model developers run out of useful data, our positive results suggest that DP synthetic data is a promising new privacy-preserving and resource-conscious data source for improving language models both big and small. 

\section*{Impact Statement}
PrE-Text relies on models that were pretrained on public data. It does not provide privacy guarantees for the public data that was used to train these models.
Although that training data was public, it may not have been intended for use in language models, which raises questions about the ethics of using the resulting models \citep{tramer2022considerations}. 
Therefore, in real deployments for PrE-Text, it is important to ensure that the public data used to pretrain the base models has been properly audited. Broadly, it is also important to communicate to users how much privacy protection they are getting when private algorithms are run on their data, and collect informed consent. 
\section*{Acknowledgments}
This work used Bridges-2 GPU at the Pittsburgh Supercomputing Center through allocation CIS240135 from the Advanced Cyberinfrastructure Coordination Ecosystem: Services \& Support (ACCESS) program, which is supported by National Science Foundation grants 2138259, 2138286, 2138307, 2137603, and 2138296 \cite{boerner2023access}.
The authors acknowledge the National Artificial Intelligence Research Resource (NAIRR) Pilot and Delta GPU for contributing to this research result.
GF and CH were supported in part by C3.ai, JP Morgan Chase, Bosch, Intel, and the Sloan Foundation. 


\bibliography{main}
\bibliographystyle{icml2024}

\newpage
\appendix
\onecolumn
\section{Additional Related Work}

\textbf{Federated Distillation.} To sidestep the problem of high communication overheads, federated distillation has been proposed as a promising approach recently \citep{NEURIPS2020_fef6f971, Abourayya2023LittleIE, Wang_2023_CVPR, Wu2021CommunicationefficientFL, NEURIPS2020_18df51b9, Cho2021PersonalizedFL}, which draw inspiration from the PATE line of work \citep{Papernot2016SemisupervisedKT, Papernot2018ScalablePL}. As far as we know, these federated distillation methods have not developed algorithms for the private federated language setting that we study. There have been works studying DP synthetic data in settings outside of text \citep{Torkzadehmahani_2019_CVPR_Workshops, Neunhoeffer2020PrivatePB}.

\textbf{Clipping strategies.} Clipping is a critical component of DP training of deep learning models, as it limits the sensitivity of the output. \citet{li2021large,bu2024automatic, bu2023differentially, kong2024unified} study how to (1) improve the computational efficiency of clipping, and (2) make it more privacy-efficient.

\textbf{Unlearning.} \citet{chen2023unlearn, jang2022knowledge, kumar2022privacy, yao2024machine} take a machine unlearning approach to remove sensitive data from a model. This approach avoids having to retrain a non-private model using private methods from scratch.

\textbf{Private inference of language models.} \citet{tang2023privacy, wu2023privacy} show how to do in-context learning with differential privacy. \citet{duan2024flocks} show how to protect the privacy of data that is used in in-context learning. \citet{du2023dp} show how to add differential privacy in the forward pass of an LLM.  \citet{mattern2022differentially} show how to anonymize data release in language models using differential privacy.

\textbf{Private dataset selection.} \citet{gu2023choosing, hou2023privately} show how to privately choose pretraining datasets. \citet{Zhou2020BypassingTA} use gradient subspaces calculated from public data to reduce the amount of noise needed for differential privacy guarantees.

\textbf{DP finetuning.} \citet{li2023privacy} investigate differentially private prompt tuning. \citet{li2022does} show conditions under which LLMs do not suffer much accuracy drop-off from using DPSGD \citep{abadi2016deep}. \citet{yu2021large} show that by using a LoRA-like reparameterization, large language models can perform differentially private training without losing much performance. \citet{nasr2023effectively} modify DPSGD to more effectively use public data.  \citet{zhang2023dpzero,tang2024private} show how to do zeroth order optimization of language models with differential privacy. \citet{yu2024privacy} show how to do DP instruction tuning. \citet{Kairouz2020DimensionII} show that DP adagrad can achieve better regret guarantees than plain DP SGD. \citet{Tramr2020DifferentiallyPL} develop strong simple baselines for DP learning of models. \citet{wang2023private} studies how to analyze overparameterized private linear regression via a theoretical lens.  \citet{yu2022individual} show how to account for privacy on an individual example level in DPSGD.  \citet{anil2021large} show how scale allows product DP finetuning of BERT. \citet{xu2021utilitarian, mireshghallah2021privacy} consider DP language models satisfying empirical notions of privacy.  

\textbf{On-device training.} \citet{Hou2021EfficientAF} study how to do saddle point optimization in the federated setting. \citet{weller2022pretrained} study the multilingual language model learning problem in federated learning. \citet{ramaswamy2020training} train production next-token-prediction models with differential privacy. \citet{xu2023training} study the private federated learning setting, modifying private on-device training to be more noise and communication efficient. \citet{gupta2022recovering} show an attack on how to recover text used to train federated text models (if differential privacy is not used). 

\textbf{Public pretraining.} \citet{ganesh2023public} perform a theoretical analysis to show why public pretraining is so critical to the deployment of DP language models. \citet{kerrigan2020differentially} show that public pretraining helps with producing DP language models.

\textbf{Best practices in DP published work.} \citet{brown2022does, zhao2022provably} studies the issue of LLM pretraining data often not explicitly designated for public use, undermining typical assumptions made about public data in DP language models.  \citet{Ponomareva2023HowTD} describe best practices when publishing DP work.

\section{Prompt}
We provide the prompt used for \texttt{expand} here.
\begin{figure}[h]
\centering
\includegraphics[width=0.5\columnwidth]{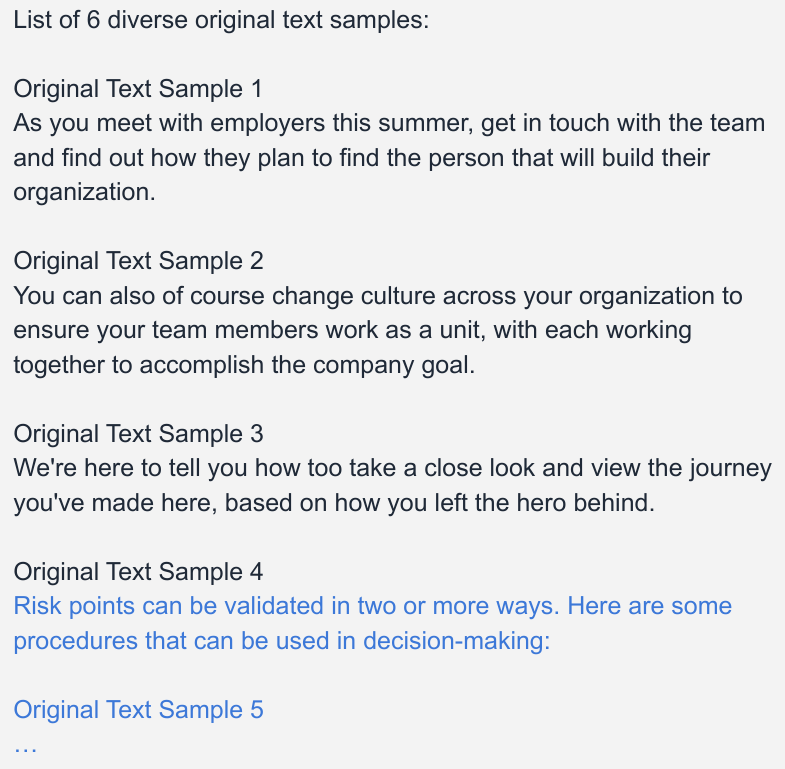}
\caption{\label{fig:prompt}The synthetic data generation prompt for \texttt{Expand}.  The blue text after ``Original Text Sample 4'' is generated.  We parse the generated text for the text between Original Text Sample 4 and Original Text Sample 5 and use that as a synthetic sample.}
\end{figure}

\section{Dataset generation details}
\label{app:dataset}
We use the train split of the c4 English (c4-en) dataset \citep{c4}. We start by producing three federated private datasets from c4-en: \textsc{Jobs}, \textsc{Forums}, and \textsc{Microblog}.  We illustrate the process for \textsc{Jobs}; the process is similar for the other two.  First, we take the first 11,000 samples in the c4-en dataset that come from a jobs site.  The private train set consists of 10,000 randomly chosen samples, and the private evaluation set consists of the remaining 1000 samples. The federated dataset is then constructed by splitting the 10,000 training samples into 1250 clients of 8 samples each, split uniformly at random. The same is done for \textsc{Forums} (from an online forum site) and \textsc{Microblog} (from an online microblogging site). 

We also evaluate our method on a question-and-answer dataset focused on coding and technical topics, which has text content partitioned by users. We construct 1250 users by making a federated dataset where each client has the comments associated with that user in the dataset. For each user, we cap the number of comments to 128. This forms the training dataset. The evaluation dataset is made up of the first 2000 samples of the next 100 users from the same dataset. We call this dataset \textsc{Code}. The results on the \textsc{Code} evaluation dataset differ slightly from the version published in ICML 2024, due to an earlier error in the train/test split, which has been corrected. The overall conclusions remain the same.

For the initial population used in \cref{private evolution}, we take random samples from c4-en that are not in the private training sets nor even from the same website sources represented in the private datasets.

\section{Experimental details}

\subsection{Privacy Details}
\label{app:privacy}
For our privacy estimate in our evaluation, we use \textsc{opacus.accountants.analysis.rdp} to compute the privacy guarantee for PrE-Text. We input $T=11$ steps, $q=1.0$, and set the \textsc{noise\_multiplier} to be the ratio of $\sigma$ to the sensitivity (the max number of samples per client for PrE-Text), setting $\sigma$ to the value that gets us the desired $\epsilon$ value).

\subsection{Baselines}
\label{app:exp details}
\textit{(1) $\epsilon=0$ baselines:} We evaluate c4-only and Expand-only. c4-only is a DistilGPT2 model finetuned on a subset of c4-en that was not in any of the private datasets. As \citet{xu2023federated} found, finetuning on c4-en improved privacy-utility tradeoff greatly in next-token prediction, so this is an important baseline. Expand-only is a DistilGPT2 model finetuned on the subset of c4-en used in the c4-only baseline expanded to 2 million samples using \texttt{expand}. We use the AdamW optimizer with a learning rate of 0.0002 and a batch size of 256 for c4-only, and a batch size of 65536 for Expand-only. We train for 20 epochs. The subset of c4-en is a subset of roughly 87k samples from c4-en that was not in any of the private datasets nor even from the same websites as any of the private datasets. We provide the precise dataset under data/initialization.json in the code repository.

\textit{(2) $\epsilon=\infty$ baselines:} We evaluate Expand-private, which is a DistilGPT2 model finetuned on (1) the subset of c4-en used in c4-only baseline, and (2) the private dataset expanded to 2 million samples using \texttt{expand}. We found that this performed better than a model finetuned on only the private dataset itself.  We use the AdamW optimizer with a learning rate of 0.0002 and a batch size of 256 for c4-only, and a batch size of 65536 and train for 20 epochs.

\textit{(3) On-device baselines:} We evaluate DP-FedAvg \citep{mcmahan2017learning} and DP-FTRL \citep{kairouz2021practical} at privacy levels of $\epsilon=1.29, \epsilon=7.58$. We first finetune DistilGPT2 on the subset of c4-en used in the c4-only baseline, and then finetune it further using DP-FedAvg or DP-FTRL (which are on-device training methods). We use the DP-FTRL-TreeRestart variant of DP-FTRL, which makes the most sense in our setting which considers full participation in each communication round. For both DP-FedAvg and DP-FTRL, we tune the client learning rate in $\{0.1, 0.01, 0.001\}$, the number of communication rounds in $\{10,20,100\}$, and clipping in $\{1.0, 0.1, 0.01, 0.001\}$. Similar to \citep{kairouz2021practical}, we found that more rounds always helped, (i.e. 100 rounds is better than 20 and 10). The batch size is 4, and server momentum is 0.9. We report the best evaluation metric, evaluated at the end of each round.

\textit{(4) Text-to-text privatization baseline:} In this approach, clients hold an LLM on-device (which may not be practical) and release privacy-preserved paraphrases of their text directly to the server. The representative method we use here is DP-Prompt \citep{utpala-etal-2023-locally}. We use the same prompt and model (flan-t5-3b) as \citet{utpala-etal-2023-locally}. Note that these methods cannot take advantage of secure aggregation (text cannot be summed together) which necessitates much more noise to be added to the privatized text. We first finetune a DistilGPT2 model on the subset of c4-en used in the c4-only baseline, and then finetune it further on the privatized text received by the server. We clip according to the top and bottom logits just like \citet{utpala-etal-2023-locally}. For the finetuning step, we use the AdamW optimizer with a learning rate of 0.0002 and a batch size of 256 and choose the best evaluation metric over 20 epochs.

\textit{(5) PrE-Text:} We use PrE-Text to generate a synthetic dataset of 2 million samples. We first finetune DistilGPT2 on the subset of c4-en used in the c4-only baseline, and then finetune it further on the synthetic dataset generated by PrE-Text. The $\Phi$ we use is miniLM-L6-v2 \citep{reimers-2019-sentence-bert}, which produces an embedding of size 384. In \texttt{Variation} we use RoBERTa-large as the mask filling model with top\_p parameter set to 1.0 and temperature set to 1.0, $W_{\text{steps}}=2$, and $\textsc{Mask}\% = 30\%$.  In \texttt{Expand} we use LLaMA-2-7B with top\_p set to 1.0 and temperature set to 1.0.  When implementing \texttt{Expand} we use the library vLLM to speed up inference \citep{kwon2023efficient}. For $\epsilon=1.29$: $N_{\text{syn}}=1024$, $H=5.9 \times 8.0 \times 1.541 \times \sqrt{2}$. For $\epsilon=7.58$: $N_{\text{syn}}=2048$, $H=8.0 \times 1.541 \times \sqrt{2}$. For \textsc{Jobs}, \textsc{Forums}, \textsc{Microblog}, there is a max of 8 samples per client, which limits the sensitivity to 8. For \textsc{Code}, we deliberately clip the number of samples per client to 16, which limits sensitivity to 16. In \textsc{Code}, both the noise and the threshold $H$ is doubled compared to the other datasets to adjust to the increased sensitivity.


\end{document}